# A Geometric Nash Approach in Tuning the Learning Rate in Q-Learning Algorithm

*Kwadwo Osei Bonsu*
*k.oseibonsu@pop.zjgsu.edu.cn*
*July 2024*

**Abstract**
This paper proposes a geometric approach for estimating the α value in Q learning. We establish a systematic framework that optimizes the α parameter, thereby enhancing learning efficiency and stability. Our results show that there is a relationship between the learning rate and the angle between a vector T (total time steps in each episode of learning) and R (the reward vector for each episode). The concept of angular bisector between vectors T and R and Nash Equilibrium provide insight into estimating α such that the algorithm minimizes losses arising from exploration-exploitation trade-off.

## 1 - Introduction

Reinforcement Learning (RL) algorithms, particularly Q-learning, are pivotal in enabling agents to learn optimal strategies through interaction with environments. Central to the effectiveness of Q-learning is the learning rate parameter α which governs how aggressively or conservatively the agent updates its state-action values based on received rewards and expected future rewards. The selection of α critically influences the algorithm's ability to balance between exploration of new actions and exploitation of known high-reward actions.

Watkins and Dayan (1992) introduced the foundational Q-learning algorithm, where α plays a key role in updating the Q-values according to the Bellman equation (Watkins & Dayan, 1992). The equation

$$Q(s,a) \leftarrow (1 - \alpha).Q(s,a) + \alpha[R_{t+1} + \gamma(\max_{a'} Q(s',a'))]$$

underscores α's influence, with $R_{t+1}$ representing the immediate reward, γ the discount factor, and $\max_{a'} Q(s',a')$ the maximum Q-value from the next state s'.

Schaul et al. (2020) uses deep Q-networks where learning is characterized by replaying transitions at the same frequency that they originally experienced, regardless of their significance; this framework which prioritizes experience, so as to replay important transitions more frequently, and therefore learn more efficiently (Schaul et al., 2020). Bhandari et al. prove finite time convergence rates for temporal difference learning with linear function approximation (Bhandari et al. 2021).

By leveraging geometric concepts such as the angular bisector between vectors T (total time steps per episode) and R (reward vector for each episode), we aim to identify an optimal α that minimizes losses attributable to the exploration-exploitation dilemma. Our approach integrates principles from Nash Equilibrium, providing a systematic framework to enhance learning efficiency and stability in RL algorithms.

## 2 - Key Definitions

1. Reinforcement Learning (RL):
   - Reinforcement Learning is a type of machine learning where an agent learns to make decisions by interacting with an environment. The agent aims to maximize cumulative rewards through trial and error learning.

2. Q-Learning:
   - Q-learning is a model-free reinforcement learning algorithm. It enables an agent to learn an optimal policy in an environment by iteratively updating a Q-value function. The Q-value represents the expected utility of taking a particular action a in a given state s.

3. State (s):
   - In reinforcement learning, a state s represents a particular configuration or situation of the environment in which the agent finds itself.

4. Action (a):
   - An action a refers to the decision or move that the agent can take in a given state s.

5. Reward (R):
   - Reward R is a scalar feedback signal that the agent receives from the environment after taking an action a in a state s. It indicates the immediate benefit or penalty associated with that action.

6. Q-Value Q(s, a):
   - The Q-value Q(s, a) represents the expected cumulative reward the agent expects to receive by taking action a in state s, and then following the optimal policy thereafter.

7. Bellman Equation:
   - The Bellman equation in Q-learning is a recursive equation that defines the optimal Q-value for a state-action pair based on the immediate reward and the discounted expected future rewards as explained by Watkins and Dayan (1992).

8. Learning Rate α:
   - The learning rate α in Q-learning determines how much the Q-value is updated each time the agent learns from new experiences. A higher α means the agent gives more weight to the most recent information, whereas a lower α makes the agent more conservative and slower to change its Q-values.

9. Discount Factor γ:
   - The discount factor γ in Q-learning represents the extent to which future rewards are discounted. It controls the importance of future rewards relative to immediate rewards. A higher γ values future rewards more, whereas a lower γ prioritizes immediate rewards. For simplicity we will not go into the complexities that arise with γ.

10. Exploration-Exploitation Trade-off:
    - Reinforcement learning agents face the dilemma of balancing exploration of unknown actions versus exploitation of known high-reward actions. Exploration allows the agent to discover potentially better strategies, while exploitation maximizes immediate rewards based on current knowledge. Learning rate α in Q-learning dictates how much the agent updates its Q-values based on new experiences. This parameter directly influences the exploration-exploitation dilemma as described below;
    - Weight of New Information: α determines the extent to which recent experiences influence Q-value updates. A higher α leads to more rapid adjustments, favoring exploration by encouraging the agent to prioritize new actions that may yield higher rewards.

    - Exploration vs. Exploitation: A high α promotes exploration as the agent quickly adapts its strategy based on recent rewards and observations. This helps in discovering potentially better policies but can lead to instability if the environment is noisy or dynamic.

    - Stability and Convergence: Optimal selection of α balances exploration and exploitation, leading to stable convergence of Q-values towards an optimal policy over time.

    - Optimizing Learning Performance: By tuning $\alpha$, we aim to maximize learning efficiency—allowing the agent to explore enough to find optimal strategies while exploiting known high-reward actions effectively.

We can therefore say that the value of α shapes how the algorithm deals the exploration-exploitation trade-off dilemmna.

3 - **Theorem 1:** Given that $\alpha, \gamma, R_t, Q \in R$ and $R_t = \frac{r_t - r_{t-1}}{r_t}$, for all $t \in N, t > 1$,

$r_t > 0$, ∃ a metric $m_1(\alpha)$ which defines a measure on T s.t. $m_1(\alpha)$ is normalized to $n_1(\alpha) = \alpha^2$ and a metric $m_2(\alpha)$ which defines $R_t$ s.t. $m_2(\alpha)$ is normalized to $0 \leq n_2(\alpha) \leq 1$.

Proof:
From the Bellman's equation $Q(s, a) \leftarrow (1 - \alpha).Q(s, a) + \alpha[R_{t+1} + \gamma(\max_{a'}Q(s', a'))]$ let's convert it into enumerated form that is, $Q_{t+1}(s, a) = (1 - \alpha).Q_t(s, a) + \alpha[R_{t+1} + \gamma(\max_{a'}Q_t(s', a'))]$.
Rearranging gives
$$\alpha = \frac{Q_{t+1}(s, a) - Q_t(s, a)}{R_{t+1} + \gamma(\max_{a'}Q_t(s', a') - Q_t(s, a))}$$
Eqn(1)

Let's define a metric $m_1(\alpha)$ that defines exploration function such that $m_1(\alpha)$ depends on the difference between Q values at time step $T_i$ and $T_{i-1}$, for $i \in N, i > 1$.

$\rightarrow \exists \varepsilon_i \in R^+$, s.t. $m_1(\alpha) = \frac{1}{N}\sum_{i=1}^{N}\varepsilon_i$

Meaning we can measure the average change in Q values for change in each time step in each episode for the entire number of episodes N. Refer to Encyclopedia of Statistics(2008) on why mean square is prefered over simple mean.

$\rightarrow m_1(\alpha) = \frac{1}{N}\sum_{i=1}^{N}\frac{1}{T_i}\sum_{t=1}^{T_i}(Q_{t+1}(s, a) - Q_t(s, a))^2$

Now we if we divide the equation by $R_{t+1} + \gamma(\max_{a'}Q_t(s', a') - Q_t(s, a))$, we end up with a normalized metric $n_1(\alpha)$

$\rightarrow n_1(\alpha) = \frac{1}{N}\sum_{i=1}^{N}\frac{1}{T_i}\sum_{t=1}^{T_i}(\frac{Q_{t+1}(s, a) - Q_t(s, a)}{R_{t+1} + \gamma(\max_{a'}Q_t(s', a') - Q_t(s, a))})^2$

comparing with equation (1) we get $n_1(\alpha) = \alpha^2$ as claimed.
Eqn(2)

Likewise since $R_t = \frac{r_t - r_{t-1}}{r_t}$ represents a percentage change in immediate rewards we can say that $0 \leq R_t \leq 1$. $m_2(\alpha)$ is a metric defining the average immediate reward for each time step in each episode for all episodes N.

$\rightarrow m_2(\alpha) = \frac{1}{N}\sum_{i=1}^{N}R_i(\alpha) = \frac{1}{N}\sum_{i=1}^{N}\frac{1}{T_i}\sum_{t=1}^{T_i}R_t$ and since $R_t$ is already normalized by expressing as a percentage it is obvious that $n_2(\alpha)$ is simply $m_2(\alpha)$.
Eqn(3)

4 - **Theorem 2:** ∃ at least one $\varepsilon$ − Nash Equilibrium between $n_1(\alpha)$ and $n_2(\alpha)$.

Proof:
4.1 - Game Setup
Our aim is to set up a two-person game scenario involving $n_1$ and $n_2$ and determine the Nash equilibrium in terms of α.

Players:
- Player 1 ($n_1$): Chooses learning rate $\alpha_i$.
- Player 2 ($n_2$): Chooses learning rate $\alpha_j$.   Where $i, j \in [1, m], \forall i, j, m \in Z^+$

Strategies:
- Each player selects a learning rate α, where α determines the rate at which the Q-values are updated during the Q-learning process.

Payoff Functions:
- Payoff for Player 1 ($n_1$):  - Represents the average reward per episode achieved when using learning rate $\alpha_i$ which Player 1 aims to maximize.

- Payoff for Player 2 ($n_2$):  - Represents convergence speed of Q-values when using learning rate $\alpha_j$ which Player 2 aims to minimize.

Nash equilibrium in this game occurs when neither player can improve their payoff by unilaterally changing their strategy (choosing a different α), given the strategy of the other player remains unchanged.

4.2 - Game Environment

1. Aligned Objectives:  Both metrics are designed to achieve complementary objectives in reinforcement learning tasks. This alignment ensures that agents prioritize actions that lead to desirable outcomes under both metrics concurrently.

2. Uniform Environment: All players operate within a consistent environment characterized by stable dynamics, consistent rewards, and penalties. This uniformity ensures that players face equivalent challenges and opportunities, facilitating fair comparison and evaluation of their strategies.

3. Proportional Impact of Learning Rate α: Changes in α, which governs the balance between exploration and exploitation, exert proportional effects on both metrics. This proportionality ensures that adjustments in α lead to consistent changes in both metrics, enabling agents to optimize effectively across multiple dimensions of learning.

3. Achieving Nash Equilibrium: Players aim to reach Nash equilibrium, where each agent's strategy is optimal given the strategies of other player. At equilibrium, agents' decisions stabilize, and the normalized values of both metrics converge or achieve a balanced state due to players' strategic optimization across both metrics simultaneously.

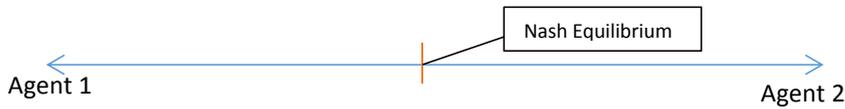

Figure 1: Line illustration of Nash Equilibrium

To determine the Nash equilibrium we need to find $\alpha^* \in \{\alpha_1, \ldots, \alpha_m\}$ such that $\alpha^* = argmin_{\alpha_i, \alpha_j} |n_1(\alpha_i) - n_2(\alpha_j)|$. Given that both metrics are continuous and differentiable in $R^+$, we set a continuous differentiable function $g(\alpha) = |n_1(\alpha_i) - n_2(\alpha_j)|$ s.t. $g(\alpha) = |p(\alpha) - q(\alpha)|$ in the same metric space.

$$\rightarrow \ g'(\alpha) = \frac{(p'(\alpha) - q'(\alpha))(p(\alpha) - q(\alpha))}{|p(\alpha) - q(\alpha)|} = 0 \ \rightarrow p'(\alpha) - q'(\alpha) = 0 \text{ or } p(\alpha) - q(\alpha) = 0$$

$$\rightarrow p'(\alpha) = q'(\alpha) \ \rightarrow \int p'(\alpha) = \int q'(\alpha) \ \rightarrow \ p(\alpha) = q(\alpha) + c, c \in R$$

This means that $\exists \ \alpha^*$ s.t. $g(\alpha^*)$ is minimum proving that there is at least one $\varepsilon -$ Nash Equilibrium between the two metrics with lower bound $\varepsilon = 0$ at $n_1(\alpha^*) = n_2(\alpha^*)$. This point also represents the point which is equidistant from both metrics in a line illustration shown in figure 1.

$$\rightarrow \ n_1(\alpha) = n_2(\alpha) = \alpha^2 = \frac{1}{N}\sum_{i=1}^{N}\frac{1}{T_i}\sum_{t=1}^{T_i} R_t \text{ is the Nash equilibrium at } \varepsilon = 0 \ .$$

Eqn(4)

A special case where Rt is a constant, we have $\alpha = \sqrt{Rt}$. This means that when we have an estimate of an unchanging R before learning, we can directly set our α as a sqaure root of R in the algorithm.

## 5 - Simulation Experiment
5.1 - Experiment Setup and Results
In order to have deeper insight into dynamisms among α, N, T and R we ran a simulation experiment and the results are shown in Figures 2 to 6. The graphs were generated with pyplot by taking in certain domains of N ∈ [1, 4000000], randomly generated values of T. We then estimate different values α and R using equation (4). Here is a list of instructions for simulation algorithm.
1- Import pyplot and relevant packages
2- Define the ranges for parameters under study
3- Generate logarithmically spaced N values from a subset of [1, 4 x 10^6]
4- Initialize arrays to store data
5- Define a function to calculate alpha based on equation (4)
6- Iterate through parameter ranges
7- Convert lists to arrays for plotting
8- Find the index of the maximum, minimum, and median average Rt
9- Print the combinations of N, Ti, and α that gives the maximum, minimum, and median average Rt
10- Plot 3D graph with colored markers
11- Add markers for the best combinations
12- Save plots as image files and provide a link to export the images
13- Print simulation results in text format

Figure 2 consists of three plots, 2D Color Graph of α vs N and T (Top Left), 3D (Top Right), Rt vs N and T (Bottom). These allow us to analyze the relationship between three sets of the four parameters under study.
Figures 3 to 6 shows 3D graphs of the parameters with Rt being the focus variable. The graphs show how different values of other parameters influence reward.

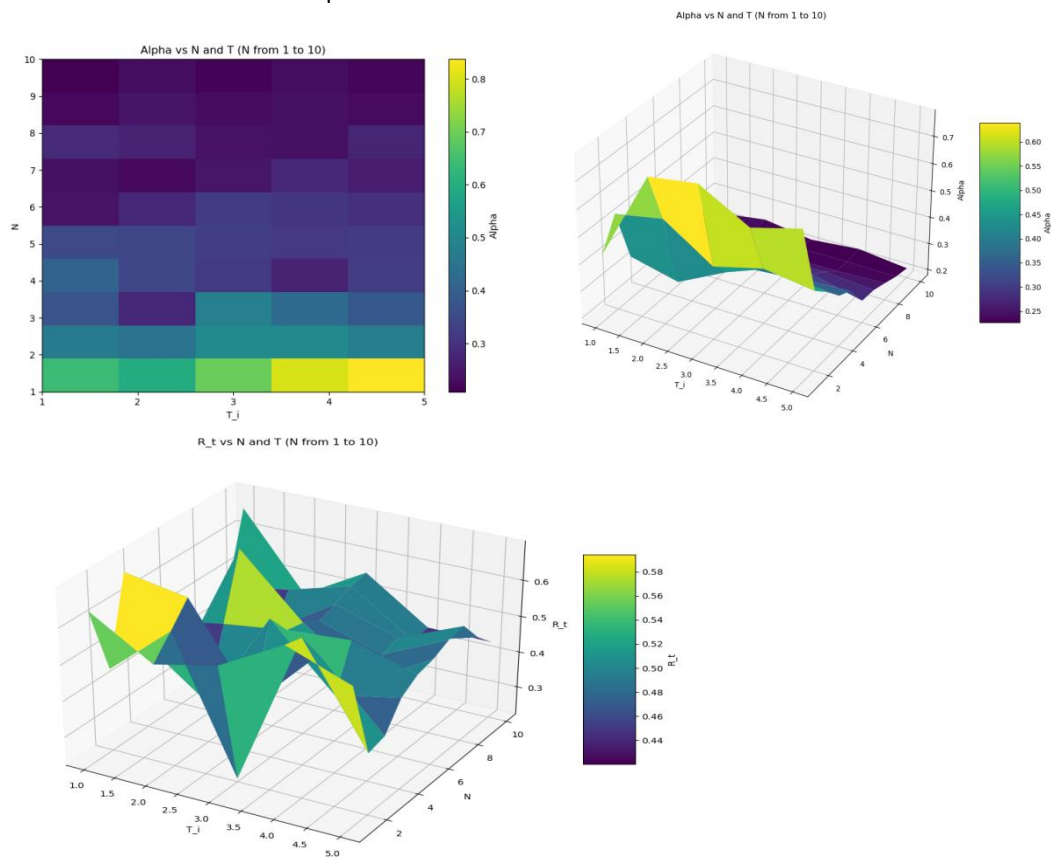

Figure 2 : 2D Color Graph of α vs N and T (Top Left), 3D (Top Right), Rt vs N and T (Bottom)

N from 1 to 10
Best combination (Max R_t): N = 3, T_i = 1, α = 0.8482513636725643, Max average R_t = 0.7195303759723649
Best combination (Min R_t): N = 1, T_i = 1, α = 0.1730807484815849, Min average R_t = 0.029956945494945653
Best combination (Median R_t): N = 5, T_i = 3, α = 0.7053332251316965, Median average R_t = 0.4974949584746805

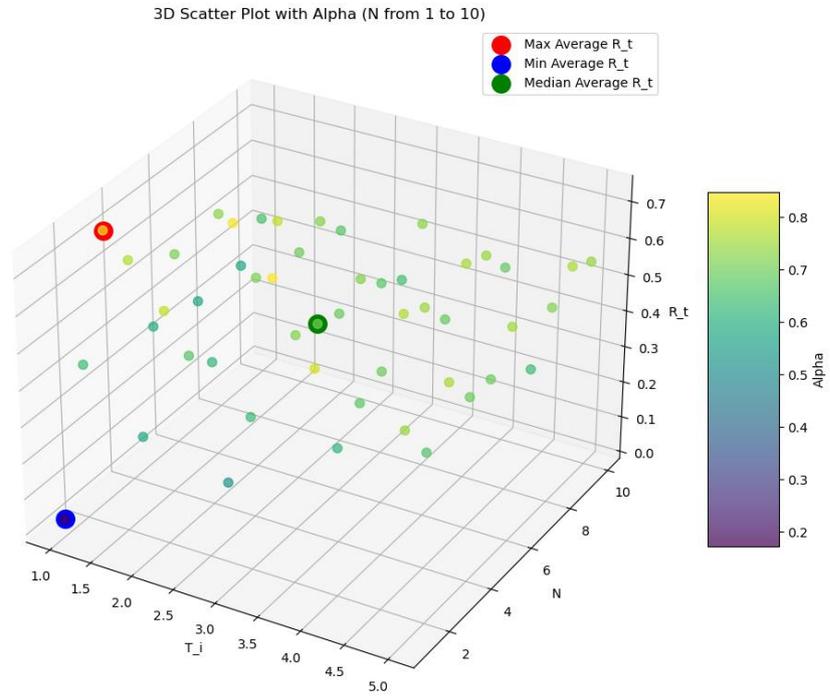

Figure 3 : 3D Graph of of Rt Vs Ti, N and α, Consider N from 1 to 10

N from 100 to 10,000
Best combination (Max R_t): N = 100, T_i = 1, α = 0.7519653589119, Max average R_t = 0.5654519010035027
Best combination (Min R_t): N = 278, T_i = 1, α = 0.6740333747090533, Min average R_t = 0.45432099022167566
Best combination (Median R_t): N = 3593, T_i = 2, α = 0.7065398969021665, Median average R_t = 0.49919862591452396

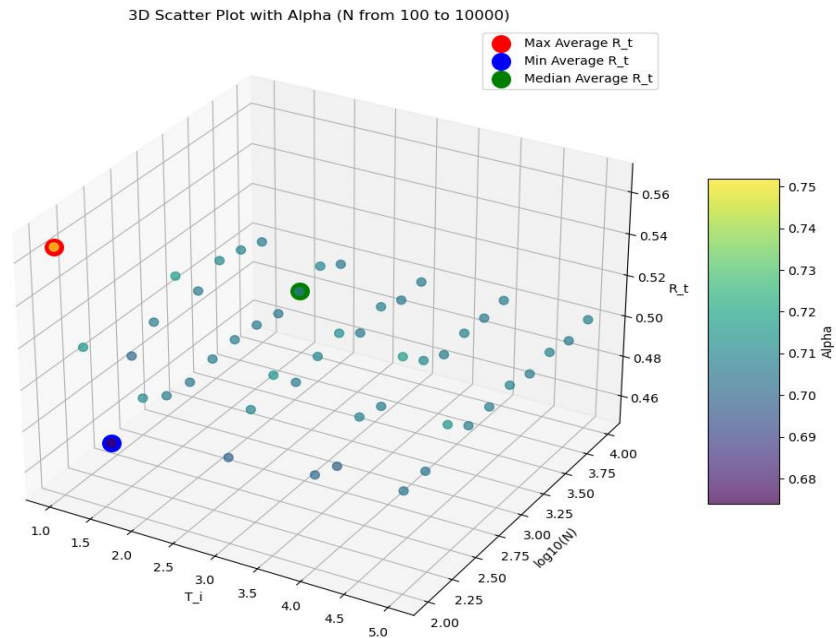

Figure 4 : 3D Graph of of Rt Vs Ti, N and α, Consider N from 100 to 10000

N from 94999 to 1 million
Best combination (Max R_t): N = 983047, T_i = 1, α = 0.7074724186504735, Max average R_t = 0.5005172231511624
Best combination (Min R_t): N = 949999, T_i = 1, α = 0.7066063093957379, Min average R_t = 0.49929247647786495
Best combination (Median R_t): N = 966382, T_i = 5, α = 0.7070954459499805, Median average R_t = 0.499983969683201

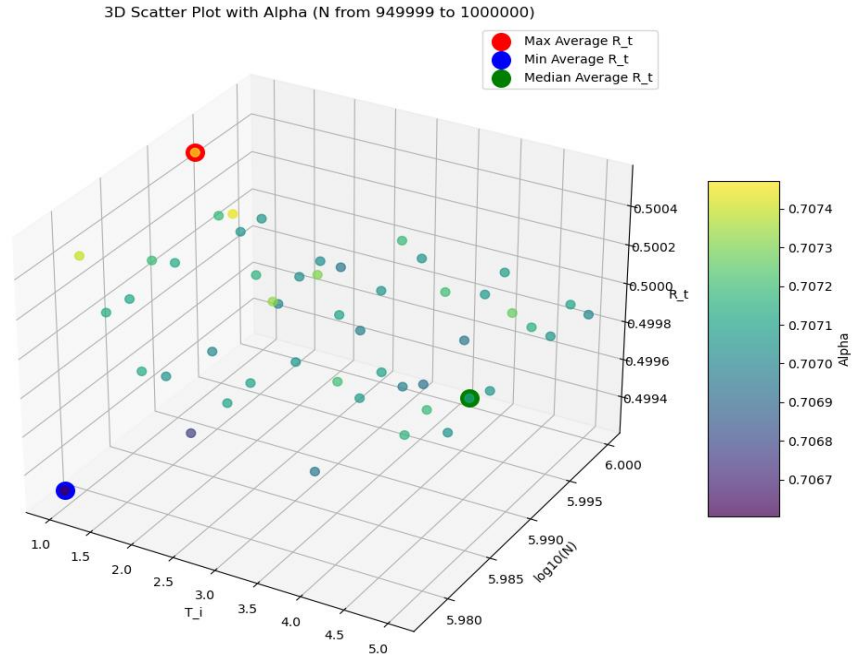

Figure 5 : 3D Graph of of Rt Vs Ti, N and α, Consider N from 94999 to 1000000

N from 100 to 4 million
Best combination (Max R_t): N = 100, T_i = 3, α = 0.7250897520627323, Max average R_t = 0.5257551485463946
Best combination (Min R_t): N = 100, T_i = 1, α = 0.6864645351276847, Min average R_t = 0.47123355798806826
Best combination (Median R_t): N = 1232310, T_i = 5, α = 0.7071883322036433, Median average R_t = 0.5001153372049751

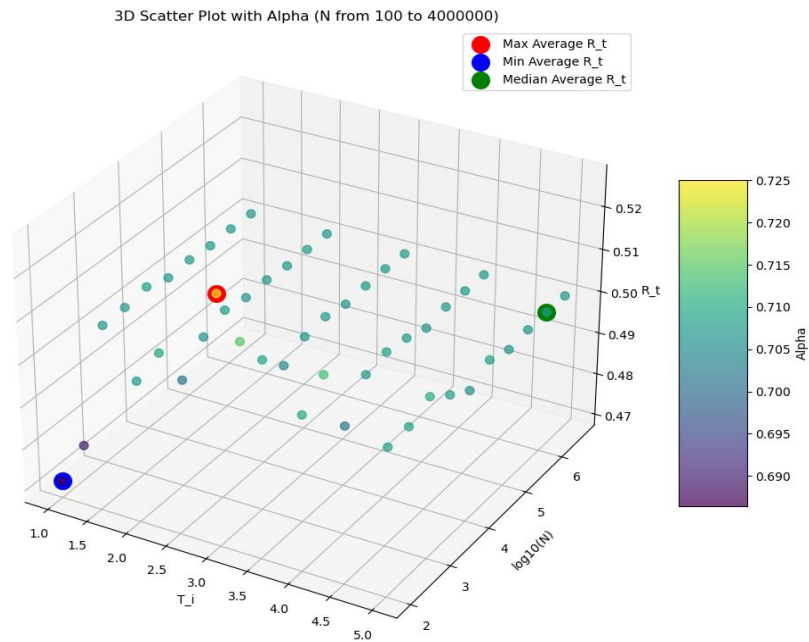

Figure 6 : 3D Graph of of Rt Vs Ti, N and α, Consider N from 100 to 4000000

5.2 - Discussion of Results
From the results obtained section 5.1 we can see that initially, within the range from N = 1 to 10, our experiment revealed significant variability in Rt outcomes. The maximum average Rt of 0.7195303759723649 was achieved at N = 3, Ti = 1, and α = 0.8482513636725643, contrasting sharply with the minimum average Rt of 0.029956945494945653 observed at N = 1, Ti = 1, and α = 0.1730807484815849. Interestingly, the median Rt was optimized at N = 5, Ti = 3, and α = 0.7053332251316965, resulting in a median average Rt of 0.4974949584746805. These results underpin the sensitivity of Rt to variations in N, Ti, and α within this initial range, highlighting the intricate dynamics of parameter optimization in RL.

Expanding our investigation to larger range N = 100 to 10,000, a clear pattern emerged where Rt values tended to stabilize as N increased. The maximum average Rt of 0.5654519010035027 was attained at N = 100, Ti = 1, and α = 0.7519653589119, while the minimum average Rt of 0.45432099022167566 occurred at N = 278, Ti = 1, and α = 0.6740333747090533. Similarly, the median Rt was found at N = 3593, Ti = 2, and α = 0.7065398969021665, resulting in a median average Rt of 0.49919862591452396. This trend towards stabilization suggests that larger sample sizes reduce variability in Rt outcomes across different parameter settings.

Extending our analysis to N from 94999 to 1,000,000 and 100 to 4,000,000, we continued to observe convergence towards an average Rt around 0.500 as N increases. Notably, α values clustered closely around 0.707 in these expansive ranges, indicating a diminishing influence of α on Rt with increasing N. This stabilization suggests a threshold where further increase in N do not significantly alter the average Rt. Our study suggests a theoretical stabilization of α around 0.707 as N becomes very large and vice versa.

6 - **Theorem 3:** There exists a homeomorphism between the discrete set T and the continuous set R as $N \to \infty$.

Corollary 1: As $N \to \infty$, the discrete set $T = \{t_1, t_2, \ldots, t_N\}$ becomes dense in the continuous interval [0, 1].
Proof: For any given $N \in Z^+$, the mapping function $f_N : T \to [0,1]$ defined by $f_N(T_i) = \frac{t_i - t_1}{t_N - t_1}$ distributes the points $t_i$ uniformly across [0, 1], where $t_1$ and $t_N$ are the minimum and maximum values in T respectively. As N increases, the spacing between consecutive points in T decreases, causing T to densely cover [0, 1] (Tareq et al. 2023).

Corollary 2: The function $f_N : T \to [0,1]$ is injective for any finite N.
Proof: The function $f_N$ maps each discrete point $t_i$ to a unique value in [0, 1], since $f_N(T_i) = \frac{t_i - t_1}{t_N - t_1}$ and distinct $t_i$ yield distinct function values. Thus, $f_N$ is an injective mapping (Munkres, 2000).

Lemma 1: The function $f_N : T \to [0,1]$ and its inverse $f_N^{-1} : [0,1] \to T$ are continuous as N becomes large.
Proof:
In the discrete topology on T, any mapping to a continuous space is trivially continuous and as $N \to \infty$, the inverse function $f_N^{-1}$ approximates a continuous mapping because the spacing between points in T becomes negligible, making $f_N^{-1}$ continuous (Escardo et al. 2001).

Lemma 2: As $N \to \infty$, the function $f_N : T \to [0,1]$ and its inverse $f_N^{-1} : [0,1] \to T$ approach a homeomorphism.
Proof: As N increases, $f_N$ becomes a bijection between T and [0, 1] with T densely covering [0, 1]. The functions $f_N$ and $f_N^{-1}$ both become close to continuous bijections, thus approximating a homeomorphism in the limit (Tareq et al. 2023).

Proof of Theorem 3: As N → ∞, the discrete set T becomes dense in the interval [0, 1] which is the range of R (Corollary 1). Consequently, as N grows, the spacing between points in T diminishes, making T increasingly dense in [0, 1] and both T and R are countably distributed within [0, 1]. This implies that for every y ∈ [0,1], ∃ $t_i$ ∈ T s.t. $f_N(T_i) \approx y$, proving that $f_N$ approaches surjectivity. The function $f_N$ is injective for any finite N (Corollary 2) since each point $t_1$ maps to a unique value in [0, 1]. Both $f_N$ and its inverse $f_N^{-1}$ which maps [0, 1] back to T, are continuous as N becomes large (Lemma 1). Knowing that $f_N$ is continuous on its corresponding discrete topology, and as N → ∞, the inverse function $f_N^{-1}$ approximates continuity due to diminishing spacing between points. This implies that as N increases, $f_N$ and its inverse $f_N^{-1}$ increasingly approximate a homeomorphism (Lemma 2).

7. **Theorem 4:** α converges around 0.707 with increasing N, given that the sets T and R are homeomorphic.

Proof:
Given that the sets T and R are homeomophic, we can project vectors T and R onto the same topological space. Let's use the same notation T and R for both the vector forms and set forms for simplicity. On such projection, symmetry can be achieved by transforming each vector onto a vector M, the angular bisector between T and R such that deviation from M will perturb the symmetry obtained. Remember, symmetry is our desired results where there is an equilibrium between exploratory behaviour and exploitatory behaviour of our Q-learning algorithm.

A geometric Nash equilibrium will be attained when both vectors T and R subtend the same angle towards M. This implies that the angle θ between T and M will be the same as the angle between R and M. Since we already proved homeomorphism between T and R, this means there must be a homeomorphism between M and R as well, so let's project M onto the normalized R space such that M inherits the properties of R.

Knowing that ∝ obtained in theorem 2 is a Nash equilibrium, consider ∝ in [0, 1] and θ in [0,π/2]. Let h: [0, 1]→ [0, π/2] be defined by h(∝) = arccos(∝). This function h establishes a continuous bijection between ∝ and θ. Hence Nash equilibrium ∝* corresponds to the same point θ* in h showing that ∝ and θ indeed share the same Nash equilibrium.

Next, by unifying equation (4) and the cosine of the angle between T and M we obtain equation (5) by squaring the angle-dot product equation of the two vectors.

$$\alpha^2 = \frac{1}{N}\sum_{i=1}^{N}\frac{1}{T_i}\sum_{t=1}^{T_i} R_t = \frac{(\sum_{i=1}^{N} T_i M_i)^2}{(\sum_{i=1}^{N} T_i^2)(\sum_{i=1}^{N} M_i^2)} = \cos^2\theta$$

Eqn(5)

WLOG, it is easy to note that the maximum angle between the angular bisector M and vectors T and R such that equilibrium is attained at $45^0$ and $\cos 45^0 = \frac{\sqrt{2}}{2}$ which is approximately 0.7071067811865475244008443 ≈ 0.707 as expected. Q.E.D.
Also the square of this ∝ value is approximately 0.5 which is the average $R_t$ value observed from simulation.

8. **Stability Analysis**
In this section we seek to find a lower upper bound on the learning rate within which a stable equilibrium is attainable.

From equation (5)
Since M has inherited properties of R including its boundaries, this implies,

$$\frac{(\sum_{i=1}^{N} T_i)^2}{(N \sum_{i=1}^{N} T_i^2)} \leq 1 \quad \rightarrow \quad (\sum_{i=1}^{N} T_i)^2 \leq N \sum_{i=1}^{N} T_i^2$$

<div align="right">Eqn(6)</div>

which is the same results as Cauchy-Schwarz inequality at $M_i = 1$, the upper bound of M.

As $N \rightarrow \infty$, we try to find the limit of the ratio of the two expressions on the inequality. We need to find a way to reduce this boundary such that the upper bound of M still satisfies. This means that the boundary found will still be able to yield high reward while minimizing the maximum exploration coefficient, thereby minimizing instability.

Equation (6) is equivalent to continuous Cauchy-Schwarz at $M_i = 1$ if we use $f_N(T_i) = \frac{t_i - t_1}{t_N - t_1}$ from theorem 3 instead of T itself so that we can compress T space into M space, thus from discrete to continuous.

$$\rightarrow \quad (\int_{t_1}^{t_N} (\frac{t - t_1}{t_N - t_1}) \, dt)^2 \leq (t_N - t_1) \int_{t_1}^{t_N} (\frac{t - t_1}{t_N - t_1})^2 \, dt$$

$$\rightarrow \quad (\int_{t_1}^{t_N} (t - t_1) \, dt)^2 \leq (t_N - t_1) \int_{t_1}^{t_N} (t - t_1)^2 \, dt$$

$$\rightarrow \quad \frac{(\int_{t_1}^{t_N} (t - t_1) \, dt)^2}{(t_N - t_1) \int_{t_1}^{t_N} (t - t_1)^2 \, dt} \leq 1 \quad \rightarrow \quad \text{LHS Ratio:} = \frac{([\frac{(t_N - t_1)}{2}]^2)^2}{(t_N - t_1) \frac{(t_N - t_1)^3}{3}}$$

$$\rightarrow \quad \text{LHS Ratio:} = \frac{\frac{(t_N - t_1)^4}{4}}{\frac{(t_N - t_1)^4}{3}} = \frac{3}{4}$$

<div align="right">Eqn(7)</div>

This obtained LHS ratio correnponds to $\alpha = \frac{\sqrt{3}}{2}$, thus $\cos 30^0$, $\therefore$ a $\theta$ of $30^0$.

Let's introduce a parameter $x \in \mathbb{R}$ such that $|\frac{t - t_1}{t_N - t_1} - x| \in [0,1]$ holds the chain of inequalities below true. This will help us determine non-inclusive boundaries for the LHS ratio within a stable equilibrium.

$$(\int_{t_1}^{t_N} (\frac{t-t_1}{t_N-t_1} - x) \, dt)^2 < (\int_{t_1}^{t_N} (\frac{t-t_1}{t_N-t_1}) \, dt)^2 \leq (t_N - t_1) \int_{t_1}^{t_N} (\frac{t-t_1}{t_N-t_1})^2 \, dt < (t_N - t_1) \int_{t_1}^{t_N} (\frac{t-t_1}{t_N-t_1} + x)^2 \, dt$$

$$\rightarrow (\int_{t_1}^{t_N} (\frac{t - t_1}{t_N - t_1} - x) \, dt)^2 < (t_N - t_1) \int_{t_1}^{t_N} (\frac{t - t_1}{t_N - t_1} + x)^2 \, dt$$

$$\rightarrow \quad \frac{(\int_{t_1}^{t_N} (t - t_1) - x(t_N - t_1) \, dt)^2}{(t_N - t_1) \int_{t_1}^{t_N} ((t - t_1) + x(t_N - t_1))^2 \, dt} < 1$$

$$\rightarrow \quad \text{LHS Ratio} = \frac{(\frac{((t_N - t_1) - x(t_N - t_1))^2}{2} - \frac{((t_1 - t_1) - x(t_N - t_1))^2}{2})^2}{(t_N - t_1)(\frac{((t_N - t_1) + x(t_N - t_1))^3}{3} - \frac{((t_1 - t_1) + x(t_N - t_1))^3}{3})}$$

$$\rightarrow \text{LHS Ratio} = \frac{3}{4}\frac{(t_N - t_1)^4((1-x)^2 - x^2)^2}{(t_N - t_1)^4((1+x)^3 - x^3)} = \frac{3}{4} \times \frac{(1-2x)^2}{1 + 3x + 3x^2 + x^3 - x^3}$$

$$\rightarrow \text{LHS Ratio} = \frac{3}{4} \times \frac{1 - 4x + 4x^2}{1 + 3x + 3x^2}$$

We see clearly see why when x=0, LHS Ratio is 3/4 as shown in equation (7).
Critical points $x = 1/2$ and $x = -7/12$ both have trivial results, thus 0 and 13 respectively.
When $x > -1/24$ we have ratio $< 1$, upholding Cauchy-Schwarz. We need to find conditions on x such that the LHS ratio is bounded between the Nash equilibrium ($\alpha^2 = 1/2$) and the upper bound 3/4.

$$\rightarrow \frac{1}{2} < \frac{3}{4} \times \frac{1 - 4x + 4x^2}{1 + 3x + 3x^2} < \frac{3}{4}, \text{ resulting in two solutions, } 0 < x < 0.0566243; \ 2.94338 < x < 7.$$

Let's focus on $0 < x < 0.0566243$ as it satifies initial conditions for $f_N(T_i)$. Based on our findings, selecting a value for x within the observed domain is crucial in tuning Q learning algorithm for effective exploration while ensuring stable reward outcomes. Our results indicate that x = 0 results in an upper bound of 3/4, whereas x = 0.0566243 corresponds to the Nash equilibrium of 1/2. Tuning parameters within certain ranges can improve both stability and efficiency significantly (Eimer et al. 2023)

9. **Conclusions**
Our research uses a geometric framework for optimizing the learning rate parameter α in Q-learning, enhancing both learning efficiency and stability. By examining the relationship between α and the angle between vectors T (total time steps per episode) and R (reward vector), we developed a method that incorporates a geometric concept such as the angular bisector and Nash Equilibrium. The experimental results reveal significant variability in reward outcomes Rt within smaller sample sizes, with the maximum and minimum rewards occurring at different α settings. Specifically, α = 0.848 yielded the highest average reward, while α = 0.173 resulted in the lowest. As sample size N increased, the results showed a trend toward stabilization of Rt around 0.5, with α converging around 0.707 for larger N.
Additionally, our analysis of the exploration-exploitation trade-off and Nash equilibrium provides further insights into balancing these elements effectively, highlighting the potential of our method in achieving stable reward outcomes.
Despite reasonable consistenty between theoretical and simulation results, there are a number of limitations which might affect direct application of our framework; for example, implementation of very large sample sizes would have provided more insight into the behaviour of the system, the effectiveness of our method may vary depending on the specific characteristics of the reinforcement learning environment. Future research should explore the application of our framework to a broader range of environments and tasks to assess its generalizability. Also, integrating our approach with advanced techniques such as deep reinforcement learning or multi-agent systems may provide new avenues for improving robustness and applicability.

**Conflicts of Interest**
There are no conflicts of interest.